\documentclass{article}

\PassOptionsToPackage{numbers, compress}{natbib}
\usepackage[final]{neurips}

\usepackage{latexsym}
\usepackage{tcolorbox}

\usepackage[T1]{fontenc}
\usepackage{inconsolata}
\usepackage[utf8]{inputenc}
\usepackage{microtype}

\usepackage{tabularray}   % 必须在导言区加载
\UseTblrLibrary{booktabs} % 让 \toprule \midrule 等可用

\usepackage{wrapfig,lipsum,booktabs}
\usepackage{pifont}
\usepackage{makecell} %multiline cell
\usepackage{tabularx}
\usepackage[normalem]{ulem}
\useunder{\uline}{\ul}{}
\usepackage[export]{adjustbox}

\usepackage{hyperref}       % hyperlinks
\usepackage{url}            % simple URL typesetting
\usepackage{booktabs}       % professional-quality tables
\usepackage{amsfonts}       % blackboard math symbols
\usepackage{nicefrac}       % compact symbols for 1/2, etc.

\usepackage{subcaption}
\usepackage{float}
\usepackage{multirow}
\usepackage{siunitx}

\usepackage{listings}
\usepackage{amsmath}
\usepackage{amssymb}
\usepackage{mathtools}
\usepackage{amsthm}
\usepackage{balance}
\usepackage{flushend}

\usepackage{lipsum}
\usepackage{graphicx}
\usepackage{subcaption}
\usepackage{tikz}
\usetikzlibrary{shapes, arrows, positioning}

\usepackage{fancyhdr}

\hypersetup{
	colorlinks,
	citecolor=gray,
	linkcolor=red,
	urlcolor=blue}

\usepackage[capitalize]{cleveref}
\crefname{section}{Sec.}{Secs.}
\crefname{table}{Table}{Tables}
\crefname{figure}{Fig.}{Figs.}
\crefname{algocf}{alg.}{algs.}
\Crefname{algocf}{Algorithm}{Algorithms}
\usepackage{tablefootnote}

\usepackage{color, colortbl,xcolor}
\usepackage{enumitem}
\usepackage{booktabs,arydshln}
\usepackage[multiple]{footmisc}

\lstdefinelanguage{json}{
    basicstyle=\ttfamily\footnotesize,
    numbers=left,
    numberstyle=\tiny\color{gray},
    stepnumber=1,
    numbersep=8pt,
    showstringspaces=false,
    breaklines=true,
    frame=single, % 单一边框
    backgroundcolor=\color{white}, % 白色背景
    stringstyle=\color{black}, % 值为深绿色
    upquote=true,
    morestring=[b]",
    literate=
     *{0}{{{\color{red}0}}}1
      {1}{{{\color{red}1}}}1
      {2}{{{\color{red}2}}}1
      {3}{{{\color{red}3}}}1
      {4}{{{\color{red}4}}}1
      {5}{{{\color{red}5}}}1
      {6}{{{\color{red}6}}}1
      {7}{{{\color{red}7}}}1
      {8}{{{\color{red}8}}}1
      {9}{{{\color{red}9}}}1,
    morecomment=[l]{//}, 
    morecomment=[s]{/*}{*/},
    commentstyle=\color{gray}\upshape,
    keywordstyle=\color{black},
    morekeywords={false,true,null}
}

\definecolor{lightred}{rgb}{1, 0.8, 0.8}
\definecolor{lightorange}{rgb}{1, 0.9, 0.8}
\definecolor{lightgray}{rgb}{0.9, 0.9, 0.9}

\newcommand{\maxval}[1]{\cellcolor{lightred}#1}
\newcommand{\secondmaxval}[1]{\cellcolor{lightorange}#1}
\usepackage{xspace}
\usepackage{natbib}
\usepackage{doi}

\newcommand{\DataMini}{{\texttt{PIN-14M}}\xspace}
\newcommand{\DataFull}{{\texttt{PIN-200M}}\xspace}

\newcommand{\nbf}[1]{{\noindent \textbf{#1}}}

\def\paperTitle{\includegraphics[height=14pt]{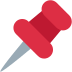} PIN: A Knowledge-Intensive Dataset for Paired and Interleaved Multimodal Documents}

\title{\paperTitle}

\makeatletter
\def\adl@drawiv#1#2#3{%
	\hskip.5\tabcolsep
	\xleaders#3{#2.5\@tempdimb #1{1}#2.5\@tempdimb}%
	#2\z@ plus1fil minus1fil\relax
	\hskip.5\tabcolsep}
\newcommand{\cdashlinelr}[1]{%
	\noalign{\vskip\aboverulesep
		\global\let\@dashdrawstore\adl@draw
		\global\let\adl@draw\adl@drawiv}
	\cdashline{#1}
	\noalign{\global\let\adl@draw\@dashdrawstore
		\vskip\belowrulesep}}
\makeatother

\setlist[itemize]{align=parleft,left=0pt..0.5em}
\setlist[enumerate]{align=parleft,left=0pt..1em}

\setlist[itemize]{align=parleft,left=0pt..0.8em}

\usepackage{booktabs}
\newcolumntype{g}{>{\columncolor{airforceblue}}c}
\def\authorBlock{
Open-source Community:\includegraphics[height=10pt]{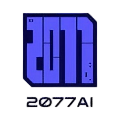} 2077AI, \includegraphics[height=10pt]{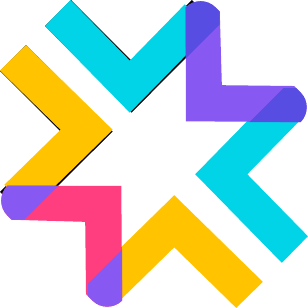} M-A-P\\

\texttt{wangjunjie@sz.tsinghua.edu.cn}, 
\texttt{ryuanab@connect.ust.hk}\\
\texttt{chenbei@01.ai}, 
\texttt{wenhu.chen@uwaterloo.ca}\\ \\

\includegraphics[height=10pt]{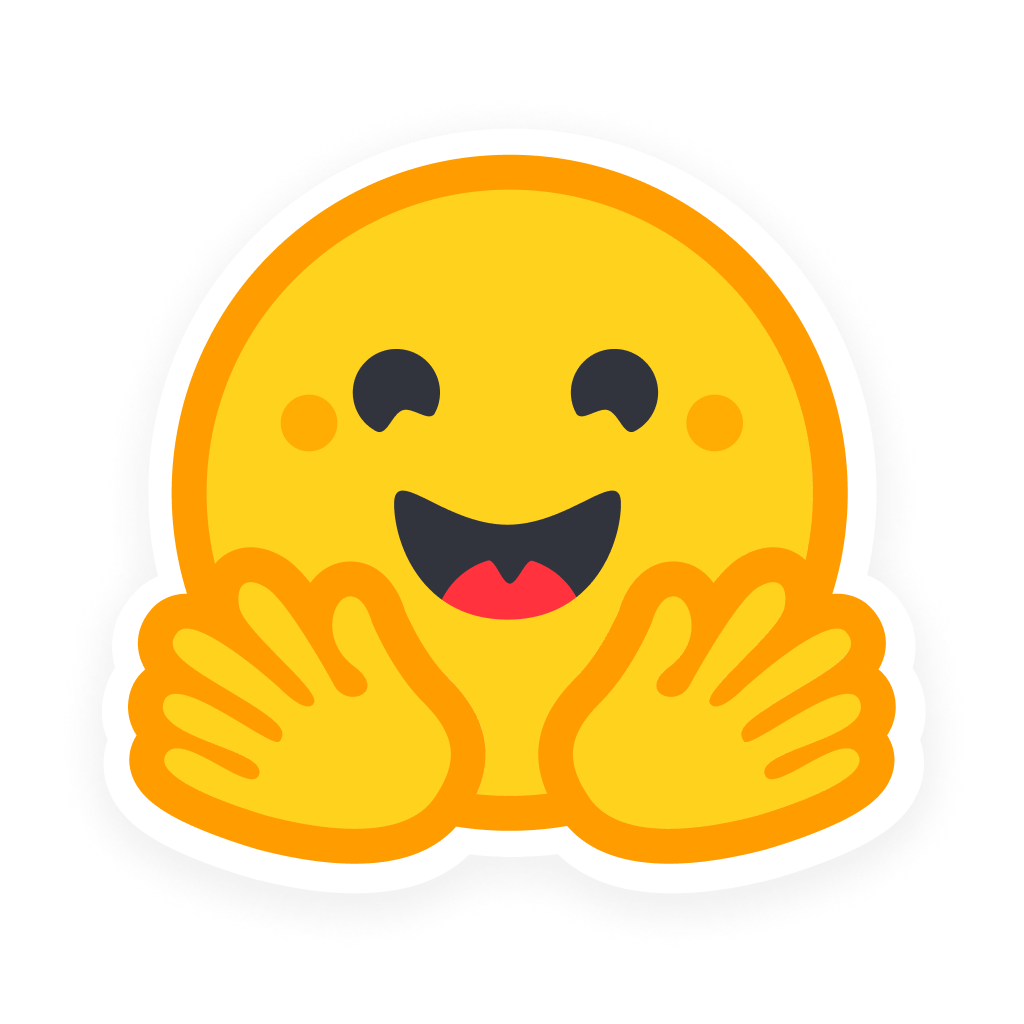} \DataMini dataset: \url{https://huggingface.co/datasets/m-a-p/PIN-14M}\\
\includegraphics[height=10pt]{figures/hf-logo.png} \DataFull dataset: \url{https://huggingface.co/datasets/m-a-p/PIN-200M}
}

\author{\authorBlock}

\begin{document}

\maketitle

% {
%   \renewcommand{\thefootnote}%
%   {\fnsymbol{footnote}}
%   \footnotetext[1]{Equal contribution.}
%   \footnotetext[2]{Corresponding Author.}
% }

\begin{abstract}
Recent advancements in large multimodal models (LMMs) have leveraged extensive multimodal datasets to enhance capabilities in complex knowledge-driven tasks.
However, persistent challenges in perceptual and reasoning errors limit their efficacy, particularly in interpreting intricate visual data and deducing multimodal relationships. 
To address these issues, we introduce PIN (\textbf{P}aired and \textbf{IN}terleaved multimodal documents), a novel data format designed to foster a deeper integration of visual and textual knowledge. 
The PIN format uniquely combines semantically rich Markdown files, which preserve fine-grained textual structures, with holistic overall images that capture the complete document layout. 
Following this format, we construct and release two large-scale, open-source datasets: \DataFull (\textasciitilde $200$ million documents) and \DataMini (\textasciitilde $14$ million), compiled from diverse web and scientific sources in both English and Chinese. 
To maximize usability, we provide detailed statistical analyses and equip the datasets with quality signals, enabling researchers to easily filter and select data for specific tasks. 
Our work provides the community with a versatile data format and substantial resources, offering a foundation for new research in pre-training strategies and the development of more powerful knowledge-intensive LMMs.
\end{abstract}

\section{Introduction}

Recent advances in large multimodal models (LMMs) have enabled their successful applications in a variety of knowledge-driven tasks such as chart reasoning and phenomenon understanding through the learning of large-scale multimodal datasets~\cite{DBLP:conf/nips/AlayracDLMBHLMM22@flamingo,DBLP:journals/corr/abs-2403-04652@yi-vl}. 
However, recent benchmark studies~\cite{DBLP:journals/corr/abs-2311-16502@mmmu, DBLP:journals/corr/abs-2401-11944@cmmmu} have highlighted two primary types of errors: perceptual errors and reasoning errors. 
Perceptual errors include difficulties in interpreting tables and graphs, especially those that are professionally complex. 
Moreover, reasoning errors often occur when the model fails to deduce relationships between images and text, particularly in scenarios involving sequential states.
In response to these challenges and with the goal of training a knowledge-intensive LMM, we adopt a data-centric solution and construct \textit{a knowledge-intensive multimodal dataset}.

\begin{figure}
\centering
\includegraphics[width=\textwidth]{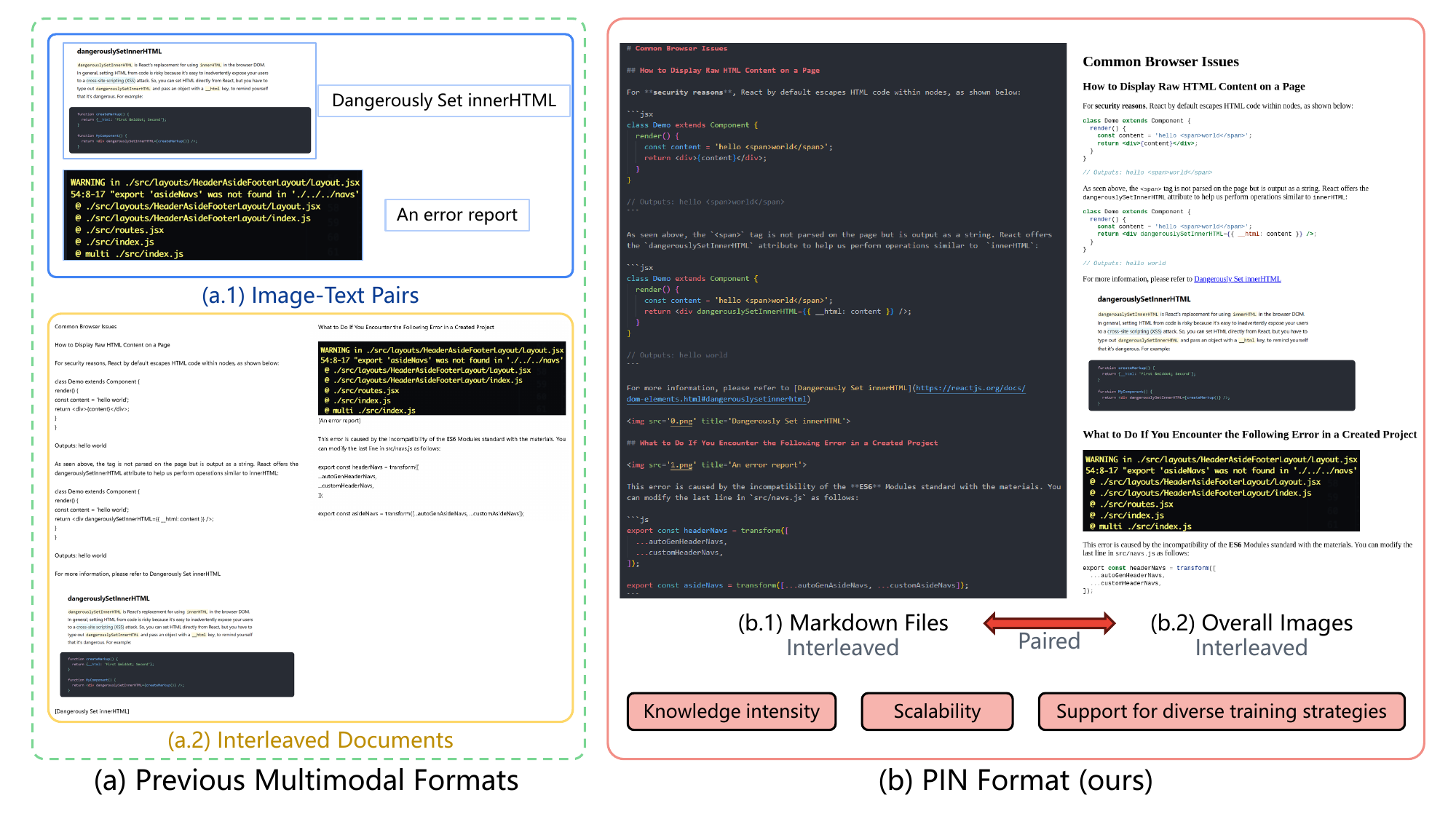}
\caption{Comparisons of traditional multimodal formats with the proposed PIN format. The PIN format preserves rich knowledge attributes (e.g., bold text, highlighting, code blocks), supports semantic interaction between images and text within Markdown documents, and enhances knowledge representation through an overall image.}
\label{fig:intro-pre-vs-pin}
\end{figure}

These datasets achieve strong results on scientific benchmarks, yet they present notable limitations. 
The exclusion of figure content or related visual elements from the text stream weakens the interaction between text and visuals within papers. 
Page-level segmentation disrupts natural continuity and impedes learning of global, document-level knowledge. The lack of open-source datasets also hinders replication.

As shown in~\cref{fig:intro-pre-vs-pin} (a), mainstream multimodal training datasets primarily fall into two categories: (a.1) image-text pairs and (a.2) interleaved documents. 
In image-text pairs~\cite{github@coyo, DBLP:conf/nips/SchuhmannBVGWCC22@laion5b}, the text corresponds to the image, allowing models to train perceptual abilities, although with limited inferential knowledge. 
Several studies shift focus to academic documents and treat paper content as text and pages as images to construct text-image datasets~\cite{DBLP:journals/corr/abs-2308-13418@nougat, DBLP:journals/corr/abs-2309-11419@kosmos-2.5}. 
These datasets achieve strong results on scientific benchmarks, yet they present notable limitations. 
The exclusion of figure content or related visual elements from the text stream weakens the interaction between text and visuals within papers. 
Furthermore, the segmentation of each page disrupts the natural continuity of the documents, impeding models from learning the comprehensive knowledge of the entire paper.
Additionally, the absence of open-source datasets presents significant challenges to replication efforts.
To capture rich interleaved information of the images and text, the interleaved document format has been introduced, enhancing both perceptual and inferential capabilities of models. 
However, this format faces three challenges: scarcity of datasets (with only MMC4~\cite{DBLP:conf/nips/ZhuHAGDFYSW023@mmc4} and OBELICS~\cite{DBLP:conf/nips/LaurenconSTBSLW23@obelics} available), a lack of specialized data (only web pages) and a lack of overall information.
We therefore seek a data format that addresses these issues. 
The ideal format should exhibit three key characteristics: \textbf{(1) knowledge intensity}, \textbf{(2) scalability}, and \textbf{(3) support for diverse training strategies}. 

Therefore, as shown in~\cref{fig:intro-pre-vs-pin} (b), we propose a novel data format: PIN, or \textbf{P}aired and \textbf{IN}terleaved multimodal documents.
Specifically, the PIN format consists of two main components: (b.1) Markdown files and (b.2) overall images. 
The first component, the Markdown files, contains knowledge-intensive, interleaved content. 
These files leverage simple markup, such as bold, italics, and headers, to explicitly define document structure and highlight key information. 
The support for embedded links and images facilitates the creation of rich multimedia documents and ensures future extensibility to additional modalities like audio and video. 
The second component is the corresponding overall image, which captures the complete visual representation of the document. 
This component enables models to learn high-level information, such as page layout, and to analyze complex relationships between text and images, including part-whole connections and sequential coherence.

This dual-component design methodically addresses three key requirements. 
First, the combination of semantically structured text from Markdown and the holistic spatial context from the overall image provides a data representation that fulfills the requirement for (1) knowledge intensity. 
Second, the PIN format is designed as a unified standard, complete with conversion processes and methods for existing datasets (as described in \cref{ss:data-process}), which ensures (2) scalability. 
Finally, this versatile structure is compatible with popular training strategies, such as image-text pair training, and also facilitates the development of novel pre-training tasks, such as generating an image from Markdown text  (as described in \cref{s:training-strategies}). 
This adaptability satisfies the requirement for (3) supporting diverse training strategies.

Following the PIN format, we release two large-scale open-source datasets: \DataFull and \DataMini. 
The \DataFull dataset contains approximately $200$ million multimodal documents, while \DataMini contains $14$ million. 
These datasets are compiled from a diverse range of Chinese and English sources.
They encompass not only common web pages but also scientific documents that include complex visualizations, such as diagrams and charts, which necessitate advanced reasoning for comprehension.
To enhance data usability, we introduce quality signals for each entry in our datasets. 
These signals are derived from extensive preliminary statistical analyses, which include metrics such as image-text interleaving frequency. 
This implementation enables the research community to perform rapid and targeted filtering to select data according to specific requirements. 
Furthermore, we present a detailed analysis of the data distribution for the \DataFull dataset. 
This analysis allows researchers to select relevant subsets for their work and serves as a reference for designing similar data processing pipelines or for choosing data sources to build custom datasets.

The contributions of this technical report are:

\begin{itemize}
    \item We propose PIN, a novel paired and interleaved multimodal document format that addresses the limitations of existing data representations. By combining semantically rich Markdown files with holistic overall images, PIN preserves fine-grained knowledge attributes and captures global document context.
    \item We introduce two large-scale, open-source datasets built on the PIN format: \DataFull (\textasciitilde $200$M documents) and \DataMini (\textasciitilde $14$M documents). Sourced from diverse Chinese and English web and scientific documents, they provide a rich resource for training LMMs on complex reasoning tasks.
    \item We enhance our datasets with quality signals and a detailed statistical analysis to improve their usability for the research community. These additions allow for targeted data filtering and provide a reproducible blueprint for future data curation efforts.
    \item We present that the PIN format's versatile structure supports both conventional training methods and enables novel pre-training tasks. This opens new research avenues, such as generating entire document images from Markdown text, to advance LMM capabilities.
\end{itemize}

\section{Related Work}

\subsection{Formats of Multimodal Data}

Multimodal pre-training datasets are primarily formatted in two ways: image-text pairs and interleaved documents. 
The \textit{image-text pair} format is the most prevalent.
This approach involves gathering large volumes of data by crawling the web for images and their corresponding alt-text descriptions~\cite{DBLP:conf/nips/SchuhmannBVGWCC22@laion5b,github@coyo,DBLP:conf/cvpr/ChangpinyoSDS21@cc12m}. 
Although these datasets achieve broad coverage, they possess inherent limitations. 
The dependency on alt-text frequently results in concise and simplistic texts that provide mere snapshots of the image content, often lacking depth in contextual richness and grammatical details. 
In many instances, the alt-text consists of only a few rudimentary words and does not form a complete sentence.
To address the issue of text quality, some methods leverage academic documents, such as scientific papers, to create higher-quality pairs~\cite{DBLP:journals/corr/abs-2308-13418@nougat,DBLP:journals/corr/abs-2309-11419@kosmos-2.5}. 
In this paradigm, a PDF page serves as the image, and the text from the document serves as the corresponding description. This strategy capitalizes on the rich semantic information inherent in academic writing. 
However, this method typically overlooks discrete embedded figures within the document, thereby missing the rich interactive information between text and visuals. 
Moreover, segmenting a document into individual pages prevents the model from leveraging information that spans across multiple pages.
The \textit{interleaved document} format emerges from more recent studies~\cite{DBLP:conf/nips/LaurenconSTBSLW23@obelics,DBLP:conf/nips/ZhuHAGDFYSW023@mmc4,DBLP:conf/nips/AlayracDLMBHLMM22@flamingo}. 
For instance, models such as OpenFlamingo~\cite{DBLP:conf/nips/ZhuHAGDFYSW023@mmc4} and IDEFICS~\cite{DBLP:conf/nips/LaurenconSTBSLW23@obelics} explore this technique by interspersing images with text in their pre-training data. 
This approach aims to enhance the multimodal recognition and reasoning capabilities of LMMs.
These methodologies involve extracting related images and text from web pages and integrating them into a coherent sequence. 
However, these strategies primarily focus on web content and often neglect knowledge-intensive sources like academic papers. 
Furthermore, rigorous data cleaning processes can discard substantial contextual information and remove crucial layout cues, such as markers. 
These procedures also fail to capture fine-grained visual details, including page layout and the specific spatial positioning of information.
To address these limitations, we propose the PIN format. 
This format is designed to maximize the extraction and presentation of both visual and textual information, thereby facilitating a more comprehensive learning environment for LMMs.

% scaling law？

\subsection{Pre-training Strategies for LMMs}

The primary objective of multimodal pre-training is to instill foundational capabilities into models by leveraging the intrinsic properties of a corpus. 
Unlike unimodal data, multimodal image-text datasets inherently contain a richer set of properties. 
These properties include the alignment between images and text, the interrelations among visual elements, and the semantic continuity within text.
Prevailing pre-training strategies for LMMs are specifically designed for particular multimodal data formats. 
For instance, strategies such as contrastive learning (CL), image-text matching (ITM), masked language modeling (MLM), and masked vision modeling (MVM) are commonly applied to image-text pair datasets~\cite{DBLP:conf/cvpr/JiWGZZWZSY23@map,DBLP:conf/icml/RadfordKHRGASAM21@clip,DBLP:conf/iclr/KwonCRBBS23@MaskVLM,DBLP:conf/acl/XuYLBHXH20@e2e-vlp}. 
In contrast, interleaved datasets enable models to perform next-token prediction by processing interwoven sequences of images and text~\cite{DBLP:conf/nips/AlayracDLMBHLMM22@flamingo}.
The proposed PIN format incorporates the characteristics of both paired and interleaved data, allowing it to seamlessly support all the aforementioned training strategies. 
Moreover, in~\cref{s:training-strategies}, we discuss potential new pre-training strategies that the unique features of our dataset make possible.

\section{Dataset Curation}

This section details the specification of our proposed data format. 
We then outline the data construction pipeline, which transforms raw documents into this structured format. 
Furthermore, we present the methods for quality control and discuss the ethical considerations associated with this dataset.

\subsection{PIN format}

\subsubsection{Philosophy}

High-quality datasets are fundamental drivers of progress across scientific and engineering disciplines, enabling advances from foundational research to industrial applications. 
We try to introduce a dataset architecture designed not only to meet current technological demands but also to remain adaptable for future advancements. 
The design is guided by three core principles:

\begin{itemize}
\item Knowledge intensity
\item Scalability
\item Supports diverse training strategies
\end{itemize}

\nbf{Knowledge intensity.}
Inspired by NOUGAT~\cite{DBLP:journals/corr/abs-2308-13418@nougat}, the proposed design enhances the knowledge density of the dataset through three primary mechanisms. 
First, in contrast to datasets derived from web pages, we prioritize the extraction of multimodal information from academic documents. 
This process converts text-only Markdown files into an interleaved format that incorporates content images. 
Furthermore, we introduce an ``overall image'' for each document to recapture the rich multimodal context lost during the data construction process of NOUGAT. 
Second, to address the limited availability of academic papers, we expand the data sources to include books and code repositories, meticulously preserving their native markup structures. 
Third, for data originating from web pages, a corresponding ``overall image'' is generated to augment the visual information within the interleaved layout. 
These methods collectively ensure the dataset possesses a high degree of informational depth.

\nbf{Scalability.}
To ensure the dataset architecture supports large-scale applications, the proposed format emphasizes two features: compatibility with existing multimodal datasets and flexibility across various data formats. 
For established multimodal datasets, such as OBELICS~\cite{DBLP:conf/nips/LaurenconSTBSLW23@obelics} and MMC4~\cite{DBLP:conf/nips/ZhuHAGDFYSW023@mmc4}, we generate an ``overall image'' for each document through straightforward processing and convert the original text-based list structures into a unified interleaved markdown format. 
Similarly, for datasets based on image-text pairs, we can easily adapt them to our format using designed templates. 
The format is also engineered to handle diverse document styles, including web pages, academic papers, and PDF files. 
Moreover, the framework accommodates text-only formats, which are indispensable for training large language models at scales exemplified by datasets like RedPajama-Data-v2~\cite{together2023@redpajama2}, which contains $30$ trillion tokens. 
We will detail how we mass-produce data in our format in~\cref{ss:data-process}.

\nbf{Supports diverse training strategies.}
The dataset format is engineered for versatility to support a wide range of pre-training strategies. 
It employs a paired and interleaved structure, comprising distinct text components (markdown files) and image components (overall images). 
This fundamental division allows for the direct application of established pre-training objectives for image-text pairs. 
The interleaved nature of the text components also ensures compatibility with advanced pre-training objectives developed for models such as Flamingo~\cite{DBLP:conf/nips/AlayracDLMBHLMM22@flamingo}. 
Beyond these, the format is designed to support emergent training paradigms, including tasks like image-based knowledge extraction. 
A comprehensive discussion of these training strategies is presented in~\cref{s:training-strategies}.

\subsubsection{Paired and Interleaved Structure}

To align with our philosophy, we design a paired and interleaved structure, as depicted in~\cref{fig:intro-pre-vs-pin} (b). 
This section specifies the organization of the dataset, detailing both the overall file structure and the composition of each data entry.

\begin{figure}[htbp]
\centering
\begin{subfigure}[b]{0.45\textwidth}
\centering
\begin{lstlisting}
example_dataset/
|
|-- content_image/
|   |-- 1.png
|   |-- 2.png
|   |-- 3.png
|   ...
|-- overall_image/
|   |-- 1.png
|   |-- 2.png
|   |-- 3.png
|   ...
\-- example_dataset.jsonl
\end{lstlisting}
\caption{Structure of the example dataset.}
\label{fig:dataset1}
\end{subfigure}
\hfill
\begin{subfigure}[b]{0.45\textwidth}
\centering
\begin{lstlisting}
example_dataset/
|
|-- part00/  # The first part.
|   |-- content_image/  
|   |-- overall_image/ 
|   \-- part00.jsonl
|
|-- part01/  # The second part.
|   |-- content_image/
|   |-- overall_image/
|   \-- part01.jsonl
|
...  - More similar parts.
\end{lstlisting}
\caption{Segmented structure of the example dataset.}
\label{fig:dataset2}
\end{subfigure}
\caption{The file tree structure of an example dataset in PIN format.}
\label{fig:example_dataset}
\end{figure}

\nbf{Directory structure.}
The directory structure for the proposed dataset is illustrated in~\cref{fig:example_dataset}. Each data entry is organized into the following components:
\begin{itemize}
    \item \texttt{content\_images/}: A directory that stores all image files embedded within or referenced by the markdown content.
    \item \texttt{overall\_images/}: A directory containing rendered images that provide a complete visual representation of the source document.
    \item \texttt{example\_dataset.jsonl}: A JSONL file that contains the primary textual content along with the associated metadata for each entry.
\end{itemize}
To enhance manageability for large-scale datasets, a partitioned structure is employed, as shown in~\cref{fig:dataset2}.

\nbf{Top-level keys.} 
ach data entry in the JSONL file adheres to a defined schema, which specifies the structure for the content, metadata, and quality signals. 
A representative example of a single data entry is illustrated in~\cref{fig:example-jsonl}.
The top-level keys are designed to encapsulate three primary categories of information: the core document content (e.g., \texttt{md}, \texttt{content\_image}), comprehensive metadata (e.g., \texttt{meta}), and utility fields for data management. 
A key feature is the \texttt{quality\_signals} field, which contains computed metrics to facilitate programmatic filtering and the creation of specialized data subsets. 
The definitions for all top-level keys are provided below.

\begin{itemize}
  \item \texttt{id} (number): A globally unique identifier for the data record.
  \item \texttt{meta} (object): A container for document-level metadata. See the nested key definitions below.
  \item \texttt{license} (string): The license associated with the sample (e.g., \texttt{CC-BY-4.0}).
  \item \texttt{quality\_signals} (object): A collection of quality indicators computed from the content, detailed in~\cref{ss:quality-signals}.
  \item \texttt{md} (string): The Markdown body text, which may contain inline \texttt{<img>} tags and mixed formatting.
  \item \texttt{content\_image} (string[]): An ordered list of content-level image paths that are referenced in the body (typically mirrors the order of \texttt{<img>} tags in \texttt{md}).
  \item \texttt{overall\_image} (string or string[]): The Path(s) to page-/document-level overall image(s). These may originate from the source dataset or be generated programmatically (e.g., a rendered screenshot of a webpage).
\end{itemize}

\nbf{Nested Keys in \texttt{meta}.} The \texttt{meta} object contains fine-grained attributes regarding the source, language, and structure of the document.
\begin{itemize}
  \item \texttt{language} (string): The primary language of the documen (e.g., en, zh).
  \item \texttt{oi\_exist} (boolean):  A flag indicating whether a document-level overall image exists.
  \item \texttt{oi\_source} (string): The source of the overall image: \texttt{ori} (from the original dataset) or \texttt{compiling} (generated programmatically).
  \item \texttt{source\_dataset} (string): Identifies the data origin. The value is either the name of the original dataset for converted entries (e.g., OBELICS), or the string \texttt{source} for natively collected entries.
  \item \texttt{ori\_meta} (object or null): A snapshot of the metadata from the original dataset. The schema of this object varies by source.
  \item \texttt{doc\_id} (number or string): Document-level unique identifier to group multiple pages/slices of the same document.
  \item \texttt{page\_id} (number or null): Page index within the document (multi-page only; single-page may be \texttt{null}).
  \item \texttt{date\_download} (string, \texttt{YYYY-MM-DD}): The date when the source document was collected.
\end{itemize}

\begin{figure}[tp]
\centering
\begin{lstlisting}[language=json]
{
    "id": 1919,
    "meta": {
        "language": "en",
        "oi_exist": true,
        "oi_source": "compiling",
        "source_dataset": "example_source (e.g. OBELICS)",
        "ori_meta": {
            "document_url": "https://www.example.com/2022/02/21/example/",
            ...
            }
        },
        "doc_id": 1997,
        "page_id": 0,
        "date_download": "2024-03-01"
    },
    "license": "CC-BY-4.0",
    "quality_signals": {
        "doc_length": 100,
        ...
    },
    "content_image": [
        "content_image/1997-0.png",
        "content_image/1997-1.png"
    ],
    "md": "<img src='content_image/1997-0.png'>\n\nThis is a fake sample data line, just for show.\n\nThis is a fake sample data line, just for show.\n\n<img src='content_image/1997-1.png'>\n\nThis is a fake sample data line, just for show.",
    "overall_image": "overall_image/1997.png"
}
\end{lstlisting}
\caption{An example data entry of JSONL files.}
\label{fig:example-jsonl}
\end{figure}

\begin{figure}
\centering
\includegraphics[width=\textwidth]{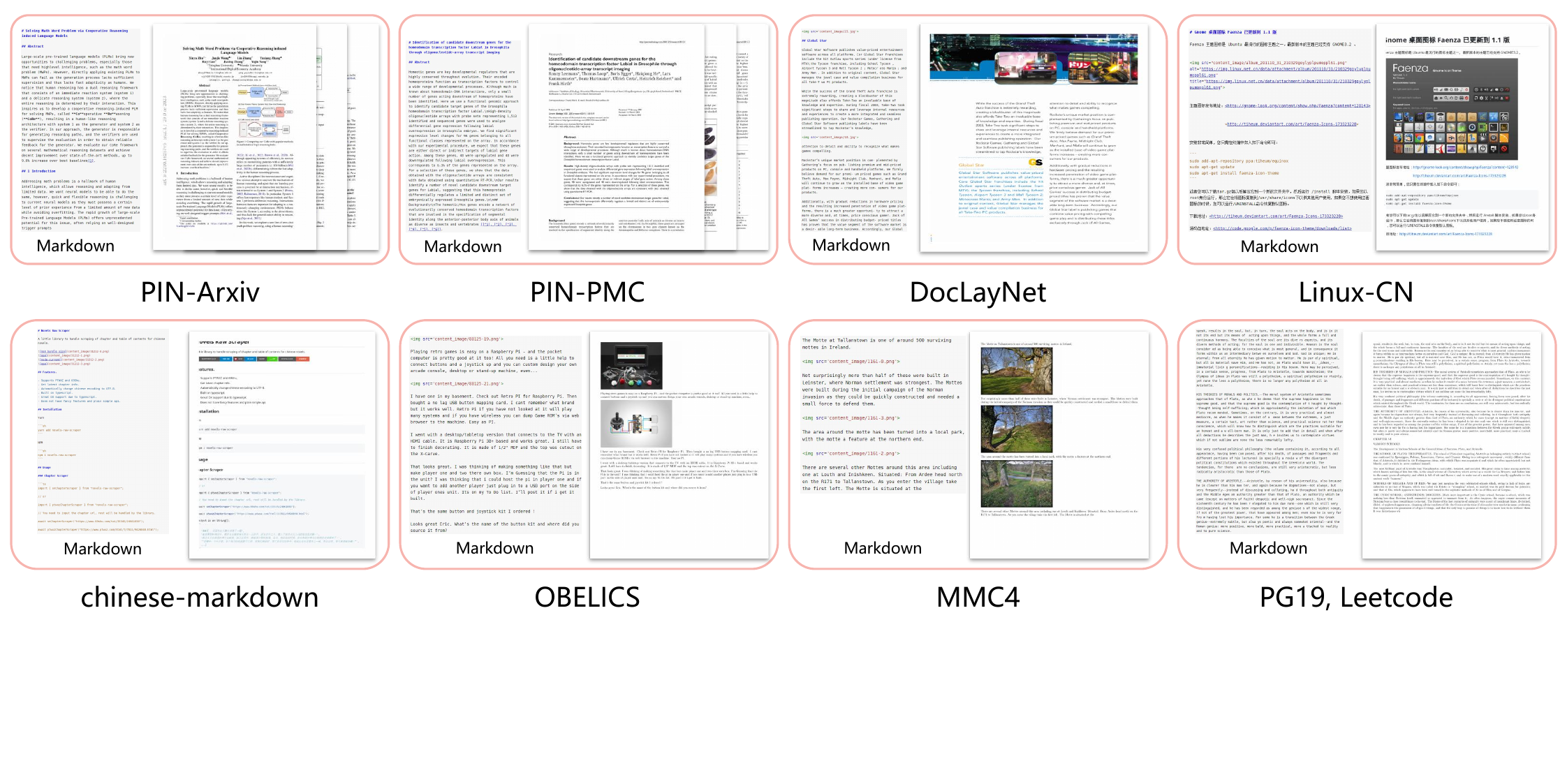}
\caption{Samples from various subsets of the \DataFull dataset. For each subset, one entry is extracted, showcasing both its markdown file section and the corresponding overall image.}
\label{fig:dataset-example}
\end{figure}

\subsection{Data Processing}
\label{ss:data-process}

\begin{figure}[tp]
\centering
\includegraphics[width=0.9\textwidth]{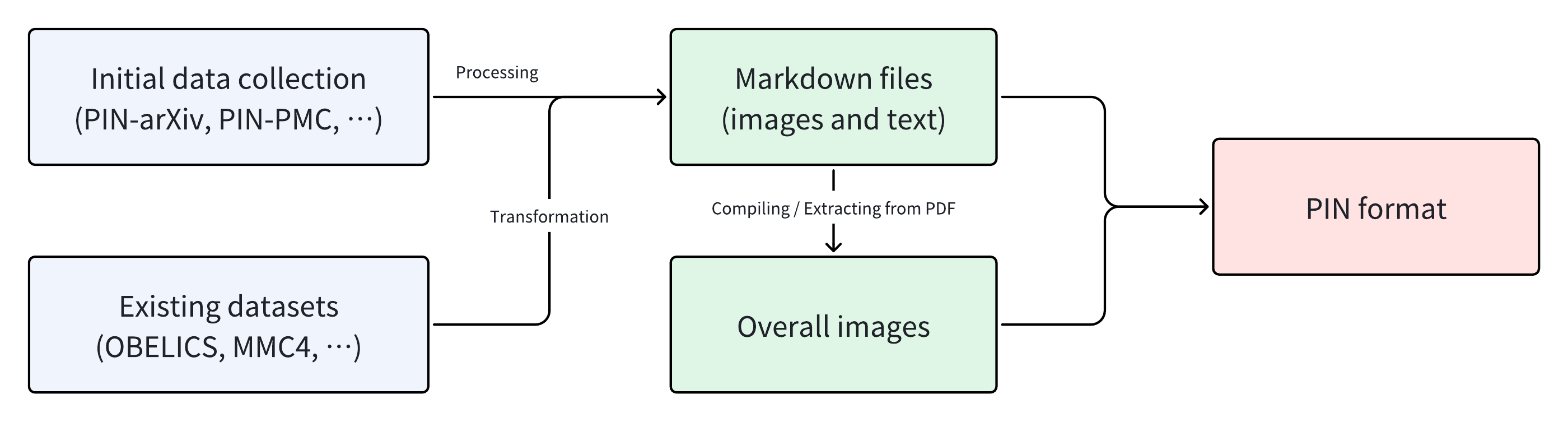}
\caption{The overview of our data process workflow.}
\label{fig:overview-process}
\end{figure}

The PIN dataset is composed of two primary categories of data: natively constructed subsets and adapted third-party datasets. 
The natively constructed subsets include PIN-PMC and PIN-Arxiv. 
The adapted datasets encompass a diverse range of sources: DocLayNet, Linux-CN, chinese-markdown, OBELICS, MMC4, Leetcode, and PG19. 
The publicly released \DataMini and \DataFull versions represent curated selections from this comprehensive collection. 
\cref{fig:dataset-example} presents a sample from each subset included in the \DataFull dataset.
As outlined in~\cref{fig:overview-process}, all data sources undergo a standardized processing pipeline to be converted into the unified PIN format. 
This pipeline consists of four main stages:
\begin{enumerate}
    \item Pre-processing: Raw text is roughly cleaned, and all associated images are downloaded and stored.
    \item Content Standardization: The text and image references are compiled into a unified Markdown file.
    \item Visual Augmentation: An ``overall image'' is generated for each entry, either by rendering the document or creating a composite image, to provide a holistic visual context.
    \item Final Assembly: The Markdown file, content images, and the overall image are packaged into the final PIN data structure.
\end{enumerate}

While the overall pipeline is consistent, specific workflows are tailored to address the unique characteristics of different data types. 
The key focus for each category is as follows:
\begin{itemize}
\item Multimodal scientific documents: Structured content from sources like PDFs or LaTeX, containing richly formatted text, figures, tables, and mathematical equations.
\item Annotated multimodal documents: Includes human-provided annotations, such as bounding boxes or layout information, alongside images and text.
\item Web pages: Features an interleaved, free-form structure of text and images derived from HTML.
\item Text-only documents: Consists purely of textual content without any associated image.
\end{itemize}

\subsubsection{Multimodal Scientific Documents}

\begin{figure}
\centering
\includegraphics[width=0.9\textwidth]{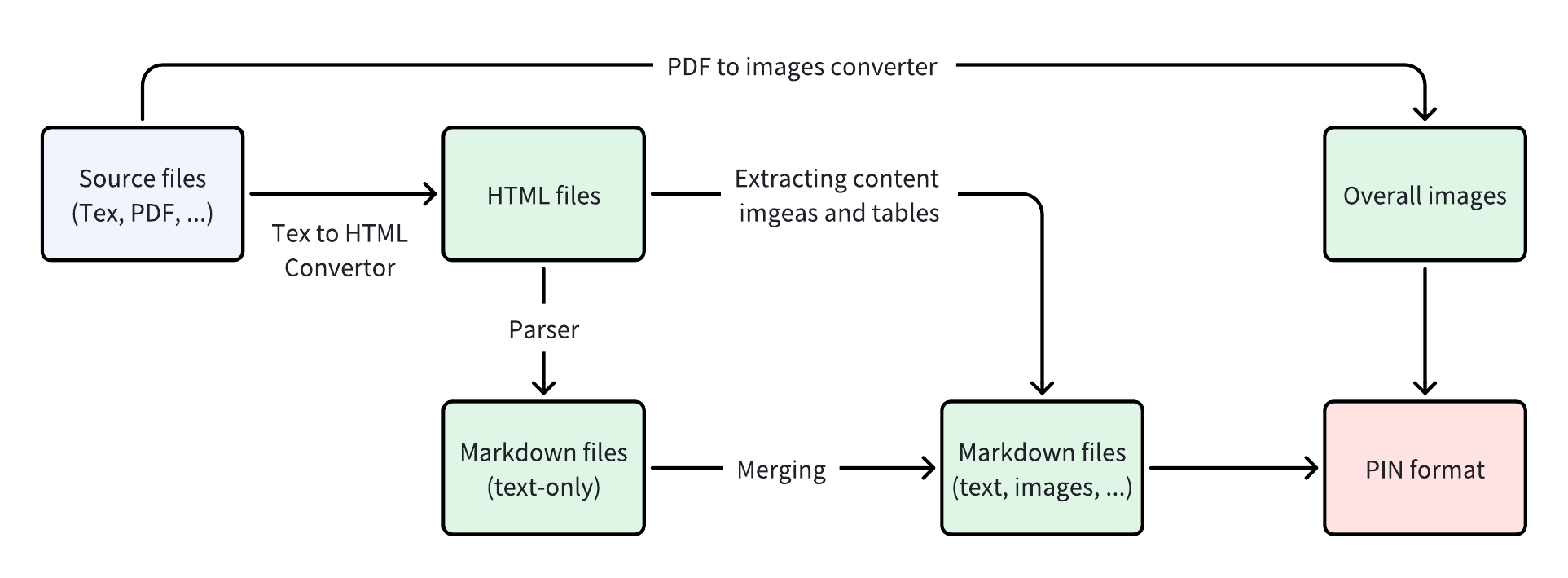}
\caption{Data processing workflow of PIN-Arxiv subset.}
\label{fig:arxiv-process}
\end{figure}

\nbf{PIN-Arxiv.}
Arxiv~\footnote{\url{https://arxiv.org/}} is a major electronic preprint platform that hosts knowledge-intensive documents across a vast range of scientific and technical fields. 
Due to this knowledge-intensive multimodal content, it serves as a primary source for the PIN-Arxiv subset. 
The data processing pipeline for this subset is depicted in~\cref{fig:arxiv-process} and consists of the following six stages:

\begin{enumerate}
\item Data Collection: The process begins by downloading the source files (primarily LaTeX) and the corresponding PDF documents from the arXiv platform.
\item Document Conversion: By utilizing the Engrafo~\footnote{\url{https://github.com/arxiv-vanity/engrafo}} converter, LaTeX documents are transformed into beautifully formatted, responsive web pages in HTML format, enhancing accessibility and visual appeal.
\item Content Parsing: The resulting HTML is then processed by the NOUGAT~\cite{DBLP:journals/corr/abs-2308-13418@nougat} parser, which extracts the textual content into a markdown format.
\item Multimodal Information Recovery: Since the initial parsing to markdown removes visual elements, a matching algorithm is employed to re-embed images and other components from the HTML source back into the markdown content, restoring the multimodal context.
\item Overall Image Generation: Each page of the original PDF documents is converted into a high-resolution image by utilizing \texttt{pdf2image} library~\footnote{\url{https://github.com/Belval/pdf2image}}.
\item Dataset Compilation: Finally, the processed Markdown file and its corresponding set of page-level ``overall images'' are assembled into the PIN format. Each data sample, therefore, represents a full document.
\end{enumerate}

This process yields the PIN-Arxiv subset, a large-scale multimodal dataset containing around $0.7$ million document samples after process around $2$ million raw papers. 
A limitation of this subset is that the content is not paginated; that is, the Markdown text is not segmented to align with individual page images. 
This is due to the significant technical challenge of reliably automating text segmentation for documents with complex layouts, such as dual-column formats. 

\begin{figure}
\centering
\includegraphics[width=0.9\textwidth]{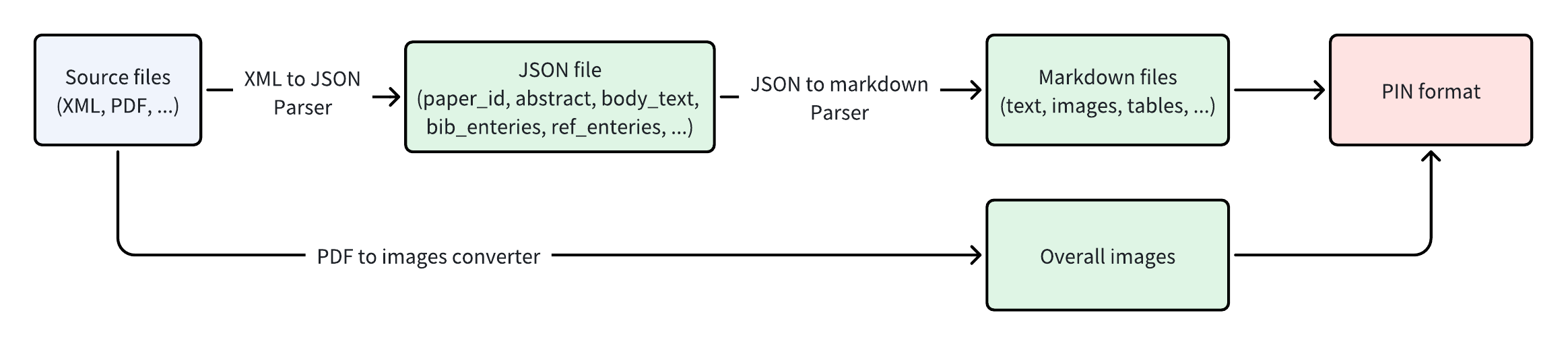}
\caption{Data processing workflow of PIN-PMC subset.}
\label{fig:pmc-process}
\end{figure}

\nbf{PIN-PMC.}
PubMed Central (PMC)~\footnote{\url{https://www.ncbi.nlm.nih.gov/pmc/}} is a free digital repository containing open-access scholarly articles in the fields of biomedicine and life sciences. 
A key feature of PMC is that its articles are available in both a human-readable PDF format and a machine-readable JATS XML format. 
Our data processing pipeline leverages the structured nature of the XML files to robustly extract core academic content while discarding purely stylistic markup.
The processing workflow, illustrated in~\cref{fig:pmc-process}, begins with the conversion of XML files to a structured JSON format. 
For this stage, we utilize a modified version of the \texttt{s2orc-doc2json} library~\footnote{\url{https://github.com/allenai/s2orc-doc2json}}. 
The original library is enhanced to improve the parsing and extraction of critical elements such as references, figures, and tables. 
Subsequently, the structured content within the JSON files is parsed and linearized into markdown documents. 
In parallel, the corresponding PDF version of each article is processed to generate a list of overall images. 
Finally, the markdown text and the associated overall images are assembled into the final PIN format.
The \DataFull dataset includes over five million samples from the PIN-PMC subset. 

\subsubsection{Annotated Multimodal Documents}

\nbf{DocLayNet.}
The DocLayNet dataset~\cite{DBLP:journals/corr/abs-2206-01062@DocLayNet} is a large-scale collection of documents featuring expert-level layout annotations. 
For each page, a corresponding JSON file provides detailed annotations, defining the bounding boxes and categorical labels (e.g., text, figure, table) for every content element.
To adapt this dataset, we develop a parser that processes these JSON annotations. 
The parser sorts the annotated content elements based on their coordinates to reconstruct the natural reading order of the document, which is then serialized into a markdown file. 
In parallel, each corresponding PDF page is extracted into an image. 
This process yields the DocLayNet subset, which contains $68,084$ samples.

\subsubsection{Web pages}

\nbf{Linux-CN.}
Technical communities and knowledge-sharing forums are valuable sources of practical, real-world expertise. 
To capture this type of information, our dataset includes content from the Linux-CN community, a prominent open-source technology forum.
The processing pipeline for this subset begins with the community's publicly available data archives~\footnote{\url{https://huggingface.co/datasets/linux-cn/archive}}. 
The initial stage involves reorganizing the raw articles and converting them into standardized GitHub Flavored Markdown (GFM) files. 
Subsequently, to generate an ``overall image'' for each article, these GFM files are rendered in a headless browser using a light theme, and a screenshot of the resulting page is captured. 
Finally, the processed markdown files and their corresponding rendered images are assembled into the PIN format. 
This process yields the Linux-CN subset, comprising $9,564$ documents.

\nbf{Chinese-markdown.}
To incorporate web-native content rich in formatted text and images, such as technical blogs and tutorials, we utilize the chinese-markdown dataset~\footnote{\url{https://huggingface.co/datasets/rojas-diego/chinese-markdown}}. 
This dataset is a collection of markdown documents curated from various web pages.
The data processing begins with a pre-processing stage where image links within the markdown files are extracted for local download, and basic text cleaning is performed. 
A key challenge with markdown documents is the difficulty of robust programmatic pagination, as elements like code blocks or tables can be improperly split across page breaks. 
To address this issue, we adopt a holistic rendering approach. 
Instead of segmenting the text, the entire markdown document is rendered as a single webpage in a headless browser using a GFM light theme. 
A full-page screenshot is then captured to serve as the ``overall image''.
Finally, the cleaned markdown documents and their corresponding full-page screenshots are assembled into the PIN format. 
This process results in the chinese-markdown subset, containing $168,323$ samples.

Usage Note: It is important to note that several samples within this subset may be flagged as potential threats by certain antivirus or security software. 
This is hypothesized to be a result of false positives, where technical articles contain inert code snippets intended as illustrative examples of security threats.

\nbf{OBELICS.}
The OBELICS dataset~\cite{DBLP:conf/nips/LaurenconSTBSLW23@obelics} is a large-scale collection of multimodal documents sourced from the web, notable for its native interleaved format of text and images. 
As the data already possesses this interleaved structure, our primary processing goal is to adapt its format and introduce page-level segmentation compatible with our schema.
A key challenge is paginating the content to generate corresponding page-level ``overall images''. 
Since the text in OBELICS generally lacks complex markup, we implement a heuristic-based pagination algorithm to segment the long-form markdown content ($\text{input}$).
This algorithm, denoted as $f_{\text{page}}$, takes a full document as input and divides it into a list of page-sized markdown segments. 
The segmentation is controlled by three key input parameters:
\begin{itemize}
\item $n_{\text{line}}$: The maximum number of lines per page.
\item $n_{\text{text}}$: The maximum number of characters per line.
\item $n_{\text{image}}$: The number of lines that an image is estimated to occupy.
\end{itemize}
This function is formally represented as:
\begin{equation}
    \texttt{page\_list} = f_{\text{page}}(\text{input}, n_{\text{line}}, n_{\text{text}}, n_{\text{image}}),
\end{equation}
where the \texttt{page\_list} consists of segmented markdown files. 
To create the visual component for each page, every markdown segment from the \texttt{page\_list} is first converted into a single-page PDF using \texttt{Pandoc}~\footnote{\url{https://pandoc.org/}}. 
This PDF is then rendered into an image file via the \texttt{pdf2image} library. 
Finally, each markdown page segment and its corresponding rendered image are assembled as a single sample in the PIN format, forming our OBELICS subset.

\nbf{MMC4.}
Similar to OBELICS, the MMC4 dataset~\cite{DBLP:conf/nips/ZhuHAGDFYSW023@mmc4} is another large-scale, interleaved multimodal dataset. 
Given its structural parallels, we adapt it to the PIN format using the same heuristic-based pagination and processing pipeline developed for the OBELICS subset. 
In detail, we process this mmc4-core-ff split to form our mmc4-core-ff subset.

\nbf{PIN-webpage (not released).}
The PIN-webpage subset is a natively constructed collection of documents crawled from various public websites. 
The entire pipeline—from data acquisition, cleaning, and filtering to the application of the heuristic-based pagination algorithm—closely follows the methodology established for processing the OBELICS dataset~\cite{DBLP:conf/nips/LaurenconSTBSLW23@obelics}. 
This subset is currently under internal development and is not included in the public data releases.

\subsubsection{Text-only documents}

\nbf{Leetcode.}
Textual data, even without visual elements, contains a wealth of structured information. To account for this, we incorporate a subset focused on richly formatted text. 
The Leetcode dataset~\footnote{\url{https://huggingface.co/datasets/greengerong/leetcode}} is selected for this purpose due to its extensive use of elements beyond plain text, such as code snippets, bolding, and underlining.
The processing pipeline begins with reorganizing the raw data into Markdown documents. 
Subsequently, we apply a rendering method similar to the one used for the Linux-CN subset to generate visual representations of these documents. 
This process yields the final Leetcode subset, which comprises $2,360$ samples.

\nbf{PG19.} 
The PG19 dataset~\cite{DBLP:conf/iclr/RaePJHL20@pg19} consists of books formatted as plain text. 
A key characteristic of this dataset is the exceptional length of the documents, which average nearly $400,000$ characters. % 399372.49 characters
To facilitate model training and the learning of pagination techniques, we segment these extensive documents into manageable, page-based units. 
Following a methodology similar to the one applied to the OBELICS subset, we first estimate the character capacity of a single page. 
Based on this estimation, each document is divided into multiple pages, with some documents spanning more than one hundred pages. 
Each resulting page and its corresponding text are treated as an individual sample. 
The final PG19 subset comprises $2,611,921$ samples.

\subsection{Quality Signals}
\label{ss:quality-signals}

Inspired by the design of the RedPajama-Data-v2 dataset~\cite{together2023@redpajama2}, we introduce a set of quality signals into the PIN format. 
Multimodal datasets often reach a massive scale, containing billions of entries. 
However, users typically have limited visibility into the intrinsic characteristics of these datasets. 
Consequently, significant data cleaning and pre-processing are necessary before such datasets can be effectively utilized for model training. 
To streamline this often repetitive process, we implement \texttt{quality\_signals} to provide researchers with a concise overview of the data characteristics, enabling rapid assessment.
In this technical report, we define the following quality signals tailored for the PIN format:

\nbf{Image-text interleaving frequency (ITIF) (\texttt{image\_text\_interleaving\_count}).} 
This metric measures the frequency of alternation between image ($I$) and text ($T$) modalities within a sequential sample. 
For a given modal sequence $S = [m_1, m_2, m_3, \ldots, m_N]$ where $m_i\in \{T, I\}$, the modality change indicator function is defined as:
\begin{equation}
\delta(i)=
\begin{cases}
1, & \text{if } m_i \ne m_{i+1},\\
0, & \text{if } m_i = m_{i+1}.
\end{cases}
\end{equation}

The ITIF is then calculated as the average number of modality changes:
\begin{equation}
\operatorname{ITIF}(S) = \frac{1}{N-1} \sum_{i=1}^{N-1} \delta(i).
\end{equation}

For instance, the sequence $T \rightarrow I \rightarrow T$ has an ITIF score of $2$, while $T \rightarrow I \rightarrow  I \rightarrow  T$ scores $2$.

\nbf{Text block count (TBC) (\texttt{text\_block\_count}).} 
This signal counts the total number of text blocks in a sample. 
A text block is a continuous segment of Markdown text, such as a paragraph, a list item, a code block, or a heading, separated by images or other modal units. 
Given a modal sequence $S = [m_1, m_2, m_3, \ldots, m_N]$ where $m_i\in \{T, I\}$, the TBC is defined as:
\begin{equation}
\operatorname{TBC}(S) = \sum_{i=1}^{N}1[m_i=T],
\end{equation}
where $1[m_i=T]$ is an indicator function that equals $1$ if the $i$-th unit is a text block and $0$ otherwise.

\nbf{Total token count (\texttt{total\_token\_count}).}
This represents the total number of tokens in the entire Markdown file, as determined by a specified tokenizer. 
In this work, we use the \texttt{meta-llama/Llama-3.2-1B} tokenizer.

\nbf{Document length (\texttt{doc\_length}).} 
This is the total character count of the entire Markdown file, calculated using the standard \texttt{len} function in Python.

\nbf{Average tokens per text block (\texttt{avg\_tokens\_per\_text\_block}).}
This metric indicates the average number of tokens contained within each text block.

\nbf{Average characters per text block (\texttt{avg\_text\_block\_length}).}
This metric represents the average character length of each text block, also determined using the \texttt{len} function.

\nbf{Markup statistics.} 
We posit that Markdown syntax contains valuable structural information. 
Therefore, we compute statistics on markup elements, including the counts of bold (\texttt{bold\_char\_count}), italic (\texttt{italic\_char\_count}), and title (\texttt{title\_count}) tags.

These signals provide an effective mechanism for identifying and filtering low-quality or irrelevant data entries.
This process minimizes data noise and enhances the overall dataset quality, a critical factor for training robust machine learning models. 
Furthermore, providing explicit quality signals enhances the transparency of the data curation process. 
These signals allow users to better understand the composition and limitations of the dataset. 
This understanding helps researchers and developers make more informed decisions regarding data selection and model training strategies.

In practice, these signals enable users to stratify or group data based on specific quality criteria. 
This capability allows users to prioritize high-quality data for training while applying distinct strategies, such as further cleaning or outright exclusion, to lower-quality subsets.

\subsection{Ethical Considerations}

Given the diverse sources of our dataset and the complex processes involved, each sample within our dataset is accompanied by a \texttt{license} field that specifies the licensing terms of the data. 
Data and components produced internally, such as compiled images, are governed by the Apache 2.0 license~\footnote{\url{https://www.apache.org/licenses/LICENSE-2.0}}.

Regarding content safety, we perform filtering to remove Not-Safe-For-Work (NSFW) images from data collected directly. 
However, given the vast scale of the dataset, exhaustive moderation is not feasible. 
Users of the data bear the responsibility to conduct further inspection and ensure compliance with all applicable laws and regulations for their specific use cases.

We welcome community feedback, questions, and suggestions. 
Public discussions on the official Hugging Face page for the dataset help us address potential issues and contribute to the continual improvement of the resource.

\section{Analysis of PIN}

In this section, we perform a preliminary analysis of our open-source \DataFull and \DataMini dataset.

\begin{figure}[!tp]
\centering
\includegraphics[width=\textwidth]{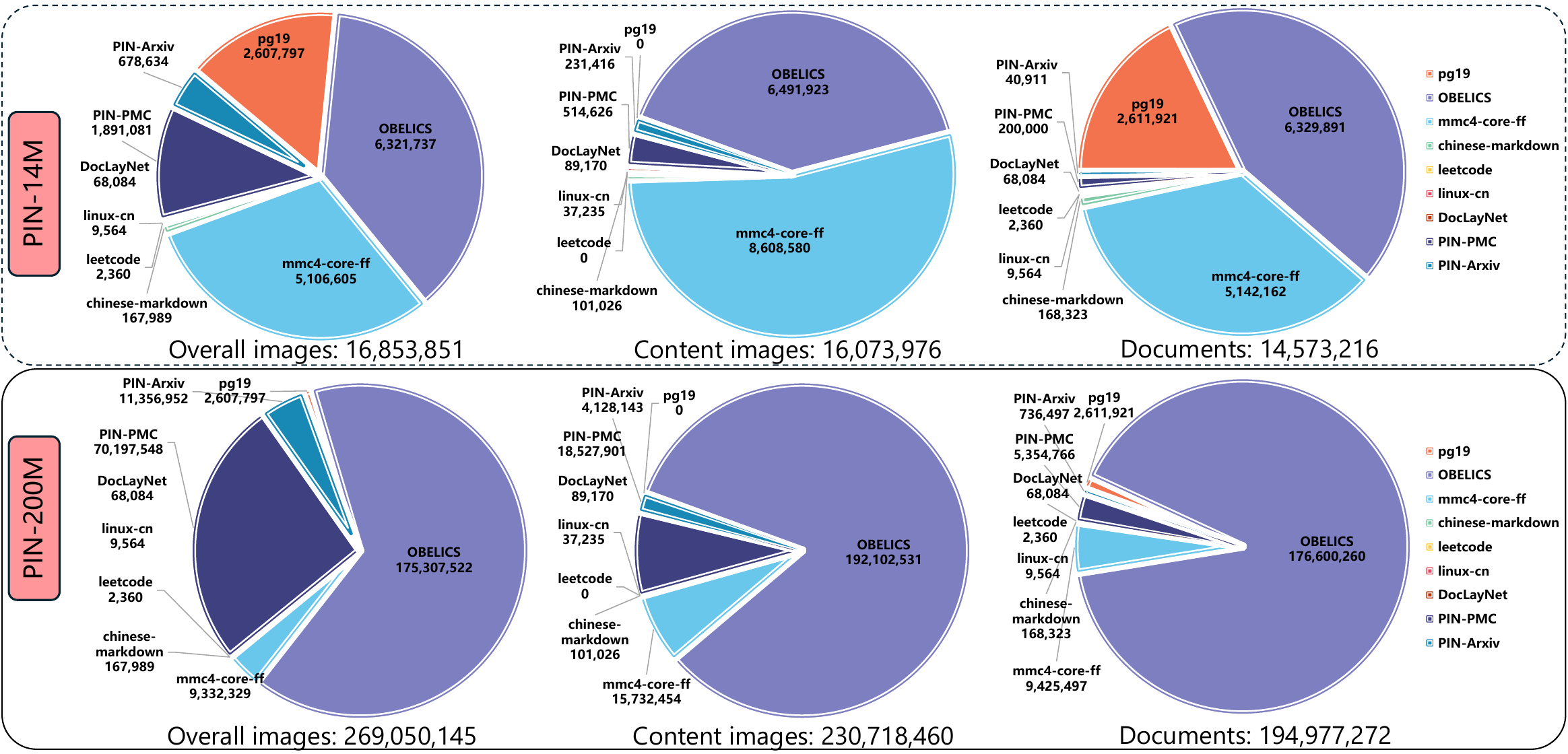}
\caption{Statistical Overview of our \DataMini and \DataFull dataset.}
\label{fig:general-stat}
\end{figure}

\subsection{Statistical Overview}

\cref{fig:general-stat} presents a quantitative breakdown of the dataset, detailing the number of documents, overall images, and content images for each subset. 
The figure also illustrates the proportional contribution of each subset to the total. 
A high-level view of the data distribution reveals that the proportions of the subsets in \DataMini are more uniform than those in \DataFull, where the OBELICS subset constitutes a significant majority.

Furthermore, a discrepancy between the document and overall image counts is apparent across several subsets, stemming from two main factors. 
First, in subsets such as PIN-PMC, the absence of pagination causes a single document to correspond to multiple overall images. 
As a result, PIN-PMC contains the second-highest number of overall images despite a relatively low document count. 
Second, some subsets employ a post-pagination process to convert markdown files into images. 
This conversion can introduce generation failures, leading to a higher number of documents than successfully created images, an effect particularly evident in the OBELICS and mmc4-core-ff subsets.

The diversity in composition and scale observed across these subsets underscores the flexibility of the PIN format, highlighting the capability for data ingestion from various sources and for straightforward large-scale integration.

\begin{table}[!tp]
\centering
\caption{Detailed signal statistics of the \DataFull{} dataset and its nine constituent subsets, highlighting the highest (\colorbox{lightred}{light red}) and second-highest (\colorbox{lightorange}{light orange}) values among subsets. The final ``\DataFull{} (total)'' row shows aggregate statistics, where the average (Avg.) is the mean of the values from the nine subsets. The ``images'' indicates content images.}
\label{tab:detailed-statistics}
\resizebox{\textwidth}{!}{%
\begin{tabular}{lrrrrrrrrrr}
\toprule
\textbf{Subset} & \textbf{\begin{tabular}[c]{@{}r@{}}Total \\ \# docs\end{tabular}} & \textbf{\begin{tabular}[c]{@{}r@{}}Total \\ \# images\end{tabular}} & \textbf{\begin{tabular}[c]{@{}r@{}}Avg. \\ \# images\end{tabular}} & \textbf{\begin{tabular}[c]{@{}r@{}}Avg. \\ ITIF\end{tabular}} & \textbf{\begin{tabular}[c]{@{}r@{}}Total \\ \# tokens\end{tabular}} & \textbf{\begin{tabular}[c]{@{}r@{}}Total \\ \# Length\end{tabular}} & \textbf{\begin{tabular}[c]{@{}r@{}}Avg. \# tokens \\ per text block\end{tabular}} & \textbf{\begin{tabular}[c]{@{}r@{}}Avg. \# \\ Bold Char.\end{tabular}} & \textbf{\begin{tabular}[c]{@{}r@{}}Avg. \# \\ Italic Char.\end{tabular}} & \textbf{\begin{tabular}[c]{@{}r@{}}Avg. \# \\ Heading Char.\end{tabular}} \\
\midrule
leetcode & 2,360 & 0 & 0.00 & 0.00 & 4.10M & 15.57M & 53.81 & \secondmaxval{13.74} & 10.18 & 5.01 \\
linux-cn & 9,564 & 37,235 & \secondmaxval{3.89} & \secondmaxval{7.20} & 17.43M & 36.27M & 46.13 & 3.02 & 12.64 & 7.00 \\
DocLayNet & 68,084 & 89,170 & 1.31 & 1.56 & 35.29M & 152.28M & 49.82 & 0.05 & 2.24 & 1.80 \\
chinese-markdown & 168,323 & 101,026 & 0.60 & 0.13 & 335.93M & 930.26M & 53.55 & 3.80 & \secondmaxval{13.77} & 8.36 \\
PIN-Arxiv & 736,497 & 4.13M & \maxval{5.61} & \maxval{8.72} & 12.10B & 39.04B & \secondmaxval{111.86} & \maxval{41.62} & \maxval{223.57} & \maxval{14.60} \\
pg19 & 2,611,921 & 0 & 0.00 & 0.00 & 2.70B & 11.50B & 69.39 & 0.01 & 3.51 & 0.01 \\
PIN-PMC & 5,354,766 & \secondmaxval{18.53M} & 3.46 & 6.05 & \secondmaxval{54.00B} & \secondmaxval{210.47B} & \maxval{123.59} & 0.68 & 5.40 & \secondmaxval{14.19} \\
mmc4-core-ff & \secondmaxval{9,425,497} & 15.73M & 1.67 & 4.01 & 2.74B & 11.97B & 105.84 & 0.003 & 0.97 & 0.0001 \\
OBELICS & \maxval{176,600,260} & \maxval{192.10M} & 1.09 & 1.46 & \maxval{72.55B} & \maxval{333.23B} & 61.96 & 0.003 & 0.44 & 0.0004 \\
\midrule
\rowcolor{lightgray}
\DataFull (total) & 194.98M & 230.72M & 1.96 & 3.24 & 144.49B & 607.35B & 75.11 & 6.99 & 30.30 & 5.65 \\
\bottomrule
\end{tabular}%
}
\end{table}

\begin{figure}[!tp]
\centering
\includegraphics[width=\textwidth]{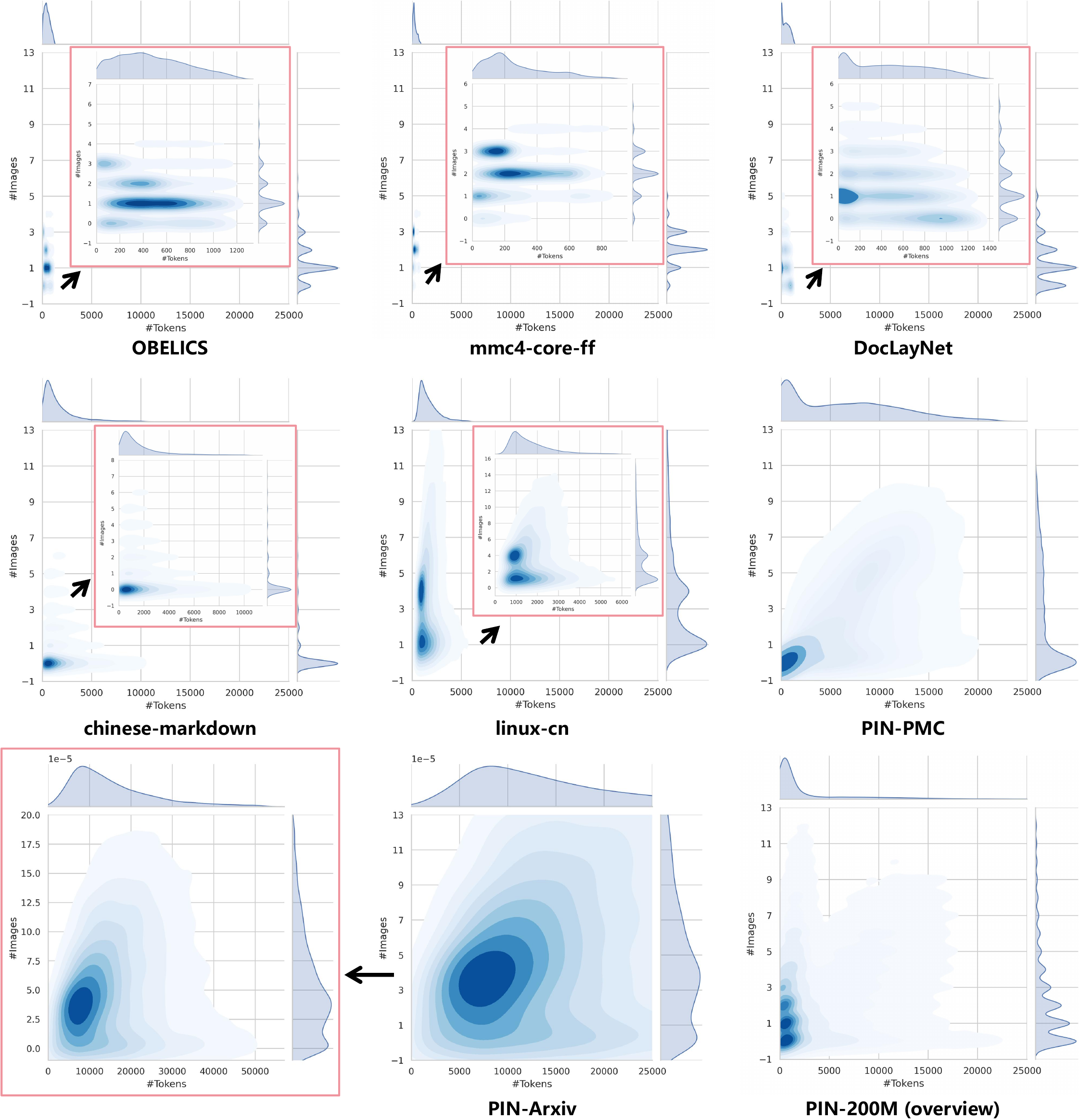}
\caption{Joint distribution of the content images and token numbers per document in \DataFull dataset.}
\label{fig:joint-dist}
\end{figure}

\subsection{Detailed Signal Statistics} 

\cref{tab:detailed-statistics} presents detailed statistics for several key quality signals across all subsets. 
These signals include the mean image-text interleaving frequency (ITIF), the average number of tokens per text block, and the prevalence of knowledge-intensive attributes (e.g., bold, italics, and headings). 
For the ``\DataFull{} (total)'' row, all average metrics are computed as the arithmetic mean of the corresponding values from the nine subsets.

Overall, the \DataFull{} dataset comprises nearly $200$ million documents, with a mean ITIF of $3.24$ and a high prevalence of knowledge-intensive attributes. 
These characteristics indicate its nature as a large-scale, knowledge-intensive resource. 
Analysis of the individual subsets reveals further findings. 
The OBELICS subset, for instance, contains the highest number of documents and tokens, as well as the greatest total length, which aligns with the finding from the previous section that it is the largest subset by proportion. 
In terms of data quality signals, PIN-Arxiv stands out. 
It exhibits the highest mean ITIF, a high average token count per text block, and the greatest prevalence of all three measured knowledge attributes, suggesting it possesses the highest density of knowledge-rich content. 
In contrast, the statistics for knowledge-intensive attributes in PIN-PMC are generally lower than those in PIN-Arxiv. 
A potential explanation is the data parsing process from XML source files, which may lead to the loss of such formatting-based attributes. 
Nevertheless, the subset remains noteworthy due to a strong ITIF and the highest average number of tokens per text block among all subsets.

In summary, these quality signals provide an effective mechanism for high-level dataset assessment and selection. 
They also enable the development of simple heuristic algorithms for fine-grained filtering at the entry level, which can significantly reduce the computational and manual effort for the research community.

\subsection{Joint Distribution Analysis}

\cref{fig:joint-dist} presents the joint distribution of the number of content images and the number of tokens per document for various subsets, alongside the overall distribution for the entire \DataFull dataset. 
This analysis is based on a random sample of $10,000$ documents from each subset that contains content images; for subsets with fewer than $10,000$ documents, the entire set is used. 
To facilitate cross-subset comparison, all distributions are visualized on a unified axis scale determined by the data range of the \DataFull (overview). 
Additionally, a view with an individually optimized scale is provided for each subset to highlight specific distributional features.

These visualizations reveal significant variations in the distributions across the subsets, underscoring the diversity of the collected data and its potential to support a wide range of applications. 
For instance, the PIN-Arxiv subset exhibits a compact and relatively uniform distribution. 
The chinese-markdown subset, in contrast, displays a distribution that is highly concentrated within a narrow range of image counts, indicating low variance in the number of images per document.

\begin{longtblr}[
  caption = {LDA results with 20 topics (trained on 100,000 sampled docs)},
  label   = {tab:topic_20},
]{width=\linewidth,  % 自动占满当前栏宽
  colspec = {p{.26\linewidth} p{.05\linewidth} p{.58\linewidth}},
  rowhead = 1,        % 让首行(表头)自动出现在每页顶部
}
\toprule
Topic Name & Ratio & Keywords\\ \midrule
Technology and Quality & 12.54 & new, system, use, used, quality, power, design, high, time, range \\
Digital Technology     &  8.60 & new, use, data, time, game, using, like, click, app, need \\
Urban Life             &  8.87 & new, first, year, two, city, home, time, team, company, years \\
Design and Aesthetics  &  8.06 & design, made, make, new, black, look, like, white, room, use \\
Politics and Society   &  7.26 & said, people, new, government, state, us, time, two, year, could \\
Entertainment and Media&  6.92 & time, like, film, new, first, love, book, life, two, show \\
General Interaction    &  5.75 & like, time, get, people, really, even, could, new, know, see \\
Health and Research    &  5.28 & water, may, many, new, research, time, health, well, used, people \\
Cooking and Recipes    &  4.34 & make, add, like, time, recipe, made, use, food, minutes, water \\
Online Activities      &  3.15 & get, new, free, online, like, use, make, time, game, best \\
Travel and Hospitality &  3.10 & park, time, new, like, hotel, get, first, take, great, said \\
Historical Events      &  2.27 & new, general, people, war, two, time, years, first, state, said \\
Daily Activities       &  1.95 & time, get, like, food, good, great, make, first, day, go \\
Personal Care          &  1.45 & skin, like, wine, time, new, love, day, first, make, get \\
Art and Museums        &  1.50 & art, work, hair, museum, first, like, time, two, new, painting \\
Cleaning and Services  &  0.88 & cleaning, car, get, services, time, new, us, said, carpet, need \\
Narratives and Dialogue&  0.84 & said, man, could, time, upon, little, like, see, two, well \\
General Opinions       &  0.55 & may, time, like, little, many, great, see, two, first, well \\
Music and Celebrations &  0.55 & music, wedding, like, said, get, time, good, make, know, bass \\
Unclassified           &  0.10 & said, time, see, like, could, make, get, little, us, well \\
\bottomrule
\end{longtblr}

\subsection{Topic Modeling}

To investigate the thematic composition of the dataset, Latent Dirichlet Allocation (LDA)~\cite{DBLP:journals/jmlr/BleiNJ03@lda} is performed on a random sample of $100,000$ documents.
\cref{tab:topic_20} presents the results of this analysis, detailing $20$ distinct topics identified by the model, along with the estimated proportion and representative terms for each topic.
The findings indicate that several major themes are prevalent, including ``Technology and Quality'', ``Digital Technology'', and ``Design and Aesthetics''. 
Beyond these dominant topics, the analysis also highlights the thematic breadth of the dataset, which encompasses a diverse array of specialized subjects such as ``Music and Celebrations'', and ``Cooking and Recipes''.

\section{Training Strategies}
\label{s:training-strategies}

\begin{figure}[h]
\centering
\includegraphics[width=\textwidth]{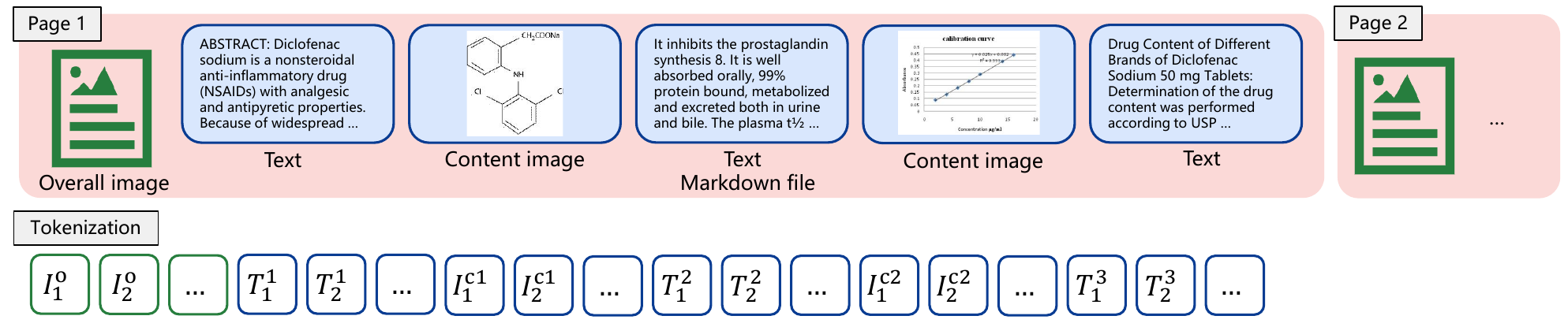}
\caption{Samples from our \DataFull dataset. Moreover, the raw data might be transformed to tokens after tokenization.}
\label{fig:raw-data-tokenization}
\end{figure}

In this section, we discuss potential training strategies, detailing how to adapt current methods and explore several new possible strategies.
The dataset comprises multimodal documents, which are structured as either single or multiple entries.
For multiple entries, \texttt{doc\_id} and \texttt{page\_id} fields facilitate coherent integration while preserving all document information.
As shown in~\cref{fig:raw-data-tokenization}, we can extract Markdown files ($S_{oi}$) and overall images ($S_{md}$) from these entries.
Following tokenization, these components are represented as:
\begin{align}
S_{oi} &=  \{ I_{i}^{o} \}_{i=1}^{|S_{oi}|} \\
S_{md} &=  \{ T_{i}^{1} \}_{i=1}^{|T^{1}|} \{ I_{i}^{c1} \}_{i=1}^{|I^{c1}|} \{ T_{i}^{2} \}_{i=1}^{|T^{2}|} \{ I_{i}^{c2} \}_{i=1}^{|I^{c2}|} \{ T_{i}^{2} \}_{i=1}^{|T^{2}|} ...
\end{align}

The training strategies discussed below are based on a conventional tokenization framework for clarity. 
It is noteworthy that this is not the sole pre-processing methodology. 
An alternative approach involves the direct processing of raw image pixels, which obviates the need for image tokenization.

\subsection{Based on Image-text Pairs}

\nbf{Contrastive learning (CL).}
The core concept is optimizing the model, allowing it to understand and align different modalities~\cite{DBLP:conf/icml/RadfordKHRGASAM21@clip,DBLP:conf/icml/JiaYXCPPLSLD21@align}. 
Specifically, corresponding image-text pairs are drawn closer in a shared embedding space, whereas non-corresponding pairs are pushed further apart. 
For instance, CLIP employs a contrastive loss function that integrates information from both image-to-text ($i2t$) and text-to-image ($t2i$) pairs. 
The specific loss function can be represented as follows:
\begin{equation}
\mathcal{L}_{i2t} = -\frac{1}{N} \sum_{i=1}^{N} \log \frac{\exp(x_i^T y_i / \sigma)}{\sum_{j=1}^{N} \exp(x_i^T y_j / \sigma)},
\end{equation}

\begin{equation}
\mathcal{L}_{t2i} = -\frac{1}{N} \sum_{i=1}^{N} \log \frac{\exp(y_i^T x_i / \sigma)}{\sum_{j=1}^{N} \exp(y_i^T x_j / \sigma)},
\end{equation}
where $N$ is the number of image-text pairs, $x_i$ and $x_j$ are the feature vectors of the $i$-th and $j$-th images, respectively, $y_i$ and $y_j$ are the feature vectors of the corresponding texts. 
Moreover, $\sigma$ is the temperature parameter. 
The $\exp$ denotes the exponential function, and $\log$ denotes the logarithm function. 
These loss functions aim to maximize the similarity between matching image-text pairs $(x_i, y_i)$ while minimizing the similarity between non-matching pairs $(x_i, y_j)$ and $(y_i, x_j)$. 
Furthermore, the overall loss is:
\begin{equation}
\mathcal{L}_{CL} = \mathcal{L}_{i2t} + \mathcal{L}_{t2i}
\end{equation}

In the PIN format, we can replace the $y$ vector with the overall multimodal vector from $S_{md}$, and the $x$ is the overall feature of $S_{oi}$. 
This enables the model to learn deeper multimodal connections by considering the relationships between overall image vectors and mixed image-text vectors. 
Moreover, when obtaining the overall vector $S_{md}$ is challenging, we can consider two sets of contrastive learning: (overall image, markup-based text) and (overall image, content images).

\nbf{Image-text matching (ITM).}
Similar to CL, ITM leverages the inherent alignment of multimodal data for pre-training~\cite{DBLP:journals/corr/abs-2003-13198@InterBERT}. 
The key difference is that ITM employs cross-entropy loss to determine whether a given image and text pair are aligned.

In the PIN format, we can use a pair of $S_{md}$ and $S_{oi}$.
Since images usually occupy a large number of tokens, we can remove the image component of $S_{md}$ to increase the difficulty of ITM task.

\nbf{Masked language modeling (MLM) and masked vision modeling (MVM).}
Both tasks involve masking some tokens and using the remaining information to reconstruct the masked portions~\cite{DBLP:conf/acl/XuYLBHXH20@e2e-vlp,DBLP:conf/emnlp/TanB19@lxmert}. 
For MLM, different segments or continuous sections of $S_{md}$ can be masked. 
For MVM, in $S_{oi}$, we can randomly mask various patches, regions, or detected objects. 
To prevent information leakage, it is essential to either synchronize or remove the image components from $S_{md}$.

\subsection{Based on Interleaved Documents}

Flamingo models the likelihood of text conditioned on interleaved sequences of text tokens and visual inputs (images/videos)~\cite{DBLP:conf/nips/AlayracDLMBHLMM22@flamingo}. 
It employs a cross-modal generation objective, which is to train the model to predict the next text token given the preceding tokens and visual context. 
The training objective can be expressed as:
\begin{equation}
    \mathcal{L}_{\text{cross-modal}} = - \sum_{t=1}^{T} \log P(w_t | w_{<t}, V),
\end{equation}
where $w_t$ represents the $t$-th token in the text sequence, and $w_{<t}$ represents all preceding tokens in the text sequence. 
$V$ represents the visual inputs (features extracted from images or videos).
In the PIN format, We can just train the models directly utilizing the interleaved part ($S_{md}$).

\subsection{Potential Strategies}

Since our format includes rich information, we might consider using only a portion of it for pre-training. 
For example, we could pre-train a robust model that understands text-rich images by focusing solely on the overall image section. 
Additionally, we could utilize the interleaved markdown file section ($S_{md}$) for the subsequent pre-training tasks such as modal prediction and multimodal next token prediction.

\nbf{Modal prediction.} 
It involves determining whether the next segment in an interleaved sequence of text and images should be text or image, based on the preceding content. 
This task leverages the known context to make accurate predictions.
A practical application involves using multimodal dialogue data, which inherently includes both text and images. 
The pre-training task focuses on predicting the content and format of subsequent dialogues.

\nbf{Multimodal next token prediction (MNTP).}
The objective is to treat all modal data, including images and text, as tokens, such as $S_{md}$. 
This approach allows the next predicted token to be either text or image, enhancing the diversity of predictions.

\nbf{Pagination prediction (PP).}
We can use the \texttt{doc\_id} and \texttt{page\_id} to determine the position of each page within the overall document. 
This allows us to assign special tokens to data subsets during pagination, thereby combining multiple pieces of data. 
For instance, a multimodal document ($S_{content}$) with two pages can be represented as follows:
\begin{equation}
S_{content} = \texttt{[BOD]} \texttt{[BOP]} S_{md}^{page 1} \texttt{[EOP]} \texttt{[BOP]} S_{md}^{page 2} \texttt{[EOP]} \texttt{[EOD]},
\end{equation}
where \texttt{[BOD]} and \texttt{[EOD]} indicate the beginning and end of the document, respectively. 
Similarly, \texttt{[BOP]} and \texttt{[EOP]} denote the beginning and end of each page. 
The PP task requires the model to predict the positions of these special tokens in conjunction with the overall images.

\nbf{Multimodal document rendering (MDR).}
This task is similar to the text-to-image generation (TIG) tasks commonly used in models like stable diffusion~\cite{DBLP:conf/cvpr/RombachBLEO22@stable-diffusion}. 
In detail, the model predicts $S_{oi}$ by learning information from $S_{md}$.
However, our situation is more challenging. 
The model not only needs to understand the text content but also to arrange the images and text appropriately. 
Additionally, it must render specific expressions of knowledge attributes, such as bold text.
We can further increase the difficulty of this task by removing all image tokens from $S_{md}$. 
This forces the model to generate suitable content images and place them in the appropriate position within the overall images.

\nbf{Knowledge extraction (KE).}
This task is analogous to image-to-text generation (ITG)~\cite{DBLP:conf/acl/XuYLBHXH20@e2e-vlp} and optical character recognition (OCR) tasks. 
ITG requires models to observe natural images and generate descriptive texts, while OCR focuses on extracting text from images along with their positional information.
In our task, the input images are text-rich article images ($S_{oi}$), and the output is the extraction of knowledge information ($S_{md}$) from these images. 
This approach ensures more natural training with reduced complexity and noise. 
Additionally, models trained using this method can seamlessly convert extensive collections of documents into interleaved multimodal formats. 
This facilitates the creation of self-iterative processes, allowing the model to generate data and continue learning autonomously.

\section{Discussions}

This work introduces a unified data format designed to seamlessly integrate diverse tasks and training processes. 
The associated data processing workflow accommodates multiple modalities and supports the straightforward incorporation of high-quality unimodal data. 
A key advantage of this uniformity is that it simplifies the analysis of scaling laws.

Furthermore, the interleaved arrangement of text and images is highly beneficial for the supervised fine-tuning phase that follows pre-training. 
This structure facilitates the direct inclusion of instructions and auxiliary information, a design choice that promotes consistency between upstream and downstream tasks. 
Consequently, the model exhibits zero-shot capabilities immediately after the pre-training stage.

The data pipelines presented in this technical report are engineered to process a wide array of document types, encompassing complex scientific articles, multimodal PDF files, standard web pages, and text-only contexts. 

We will now present some potential questions:

\textbf{Why do we not opt for OCR formats?}

The primary design objective is to enable the model to focus on high-level semantic content (knowledge), such as the meaning conveyed through images and the reasoning derived from textual information. 
This work intentionally avoids formats that rely on optical character recognition (OCR) because they introduce a significant layer of low-level perceptual details. 
Such details, including the precise positions and boundaries of individual characters, represent an unnecessary computational overhead that can divert model resources from core comprehension tasks.

For example, given an image containing the text``APPLE'', the model should ideally recognize the word as a single semantic unit. 
In contrast, an OCR-based approach would compel the model to process a complex hierarchy of character combinations (e.g., ``A'', ``AP'', ``APP'', ``APPL'', or ``APPLE'') and their corresponding spatial data. 
By abstracting away this granular level of detail, the chosen data format allows the model to allocate its resources more effectively toward knowledge understanding and logical reasoning.

\textbf{Do PIN format markdown files have a uniform style?}

GitHub Flavored Markdown (GFM) is adopted as the primary style for its widespread support across browsers and applications.
For the specific task of generating overall document images, the GFM light style is utilized. 
An exception is made for documents with a high density of mathematical formulas, such as academic papers in PIN-Arxiv subset. 
In these instances, the Mathpix Markdown format is employed due to its superior support for complex academic notations.

\textbf{How to handle tables?}

The methodology for processing tabular data is adapted according to table complexity and style. 
Tables featuring intricate or specialized designs are converted into an HTML representation to leverage the format's rich expressive capabilities and robust rendering support. 
This conversion applies, for example, to LaTeX-style tables found in academic documents. 
Conversely, simple table formats, such as those native to GFM, are maintained in their original markdown structure. 
Any table presented as a complete image is also preserved in its original format, which enables the model to learn the contextual relationships between visual tabular data and surrounding text. 
This differentiated processing strategy enhances the structural diversity of the dataset.

\section{Conclusions}

In this technical report, we introduced PIN, a novel paired and interleaved data format designed to address the persistent perceptual and reasoning limitations in Large Multimodal Models (LMMs). 
By combining semantically rich Markdown files with holistic overall images, the PIN format preserves both fine-grained knowledge attributes and global document context, overcoming the shortcomings of previous data representations. 
Based on this format, we constructed and released two large-scale, open-source datasets, \DataFull and \DataMini, derived from diverse web and scientific sources and enhanced with quality signals to improve usability. 
Our contribution provides the research community with a versatile data foundation to explore novel pre-training strategies and ultimately develop more powerful, knowledge-intensive LMMs capable of deeper multimodal understanding.

\section*{Acknowledgements}

We would like to thank Qian Liu for their insightful discussions and for providing the computational resources that supported this research.

\newpage
\section{Contributors}

\textbf{Project Leaders}
\begin{itemize}
    \item Junjie Wang; Tsinghua University, 2077AI, M-A-P, \thanks{\textcolor{blue}{\texttt{wangjunjie@sz.tsinghua.edu.cn}}}
    \item Yuxiang Zhang; Tsinghua University (Internship), 2077AI, M-A-P
    \item Minghao Liu; 2077AI, M-A-P
\end{itemize}

\textbf{Core Contributors}
\begin{itemize}
    \item Yin Zhang; Independent Researcher
    \item Yatai Ji; Tsinghua University
    \item Weihao Xuan; The University of Tokyo, 2077AI
    \item Nie Lin; The University of Tokyo, 2077AI
\end{itemize}

\textbf{Contributors}
\begin{itemize}
    \item Kang Zhu; 01.AI
    \item Zhiqiang Lin; Tsinghua University
    \item Yiming Ren; Tsinghua University
    \item Chunyang Jiang; HKUST
    \item Yiyao Yu; Tsinghua University
    \item Zekun Wang; 01.AI
    \item Tiezhen Wang; Hugging Face
\end{itemize}

\textbf{Advisors \& Corresponding Authors}
\begin{itemize}
    \item Wenhao Huang; 01.AI
    \item Jie Fu; Independent Researcher
    \item Qunshu Lin; AbakaAI
    \item Yujiu Yang; Tsinghua University
    \item Ge Zhang; University of Waterloo, 01.AI, M-A-P
    \item Ruibin Yuan; HKUST, M-A-P, \thanks{\textcolor{blue}{\texttt{ryuanab@connect.ust.hk}}}
    \item Bei Chen; 01.AI, \thanks{\textcolor{blue}{\texttt{chenbei@01.ai}}}
    \item Wenhu Chen; University of Waterloo, \thanks{\textcolor{blue}{\texttt{wenhu.chen@uwaterloo.ca}}}
\end{itemize}

{
\bibliographystyle{unsrtnat}
\bibliography{references}
}

\end{document}